\documentclass[lettersize,journal]{IEEEtran}
\usepackage{amsmath,amsfonts}
\usepackage{algorithm, algpseudocode}
\usepackage{array}
\usepackage[caption=false,font=normalsize,labelfont=sf,textfont=sf]{subfig}
\usepackage{textcomp}
\usepackage{stfloats}
\usepackage{url}
\usepackage{verbatim}
\usepackage{multirow}
\usepackage{hhline}
\usepackage{graphicx}
\usepackage{epstopdf}
\usepackage{cite}
\usepackage{hyperref}
\usepackage{makecell}

\DeclareMathOperator*{\argminA}{arg\,min}
\DeclareMathOperator*{\argmin}{min}   

\usepackage[labelfont={bf},font=small]{caption}

\newlength{\Oldarrayrulewidth}

\begin{document}

\title{Online Mutual Adaptation of Deep Depth Prediction and Visual SLAM}

\author{Shing Yan Loo$^{1,2}$, Moein Shakeri$^{1}$, Sai Hong Tang$^{2}$, Syamsiah Mashohor$^{2}$ and Hong Zhang$^{1}$
\thanks{$^{1}$The authors are with Department of Computing Science, University of Alberta, Canada.}
\thanks{$^{2}$The authors are with Faculty of Engineering, Universiti Putra Malaysia, Malaysia.}
\thanks{Manuscript received April 19, 2021; revised August 16, 2021.}}

\markboth{Journal of \LaTeX\ Class Files,~Vol.~14, No.~8, November~2021}%
{Shell \MakeLowercase{\textit{et al.}}: A Sample Article Using IEEEtran.cls for IEEE Journals}

\IEEEpubid{0000--0000/00\$00.00~\copyright~2021 IEEE}

\maketitle

\begin{abstract}
The ability of accurate depth prediction by a convolutional neural network (CNN) is a major challenge for its wide use in practical visual simultaneous localization and mapping (SLAM) applications, such as enhanced camera tracking and dense mapping.
This paper is set out to answer the following question: Can we tune a depth prediction CNN with the help of a visual SLAM algorithm even if the CNN is not trained for the current operating environment in order to benefit the SLAM performance?
To this end, we propose a novel online adaptation framework consisting of two complementary processes: a SLAM algorithm that is used to generate keyframes to fine-tune the depth prediction and another algorithm that uses the online adapted depth to improve map quality.
Once the potential noisy map points are removed, we perform global photometric bundle adjustment (BA) to improve the overall SLAM performance.
Experimental results on both benchmark datasets and a real robot in our own experimental environments show that our proposed method improves the overall SLAM accuracy.
While regularization has been shown to be effective in multi-task classification problems, we present experimental results and an ablation study to show the effectiveness of regularization in preventing \textit{catastrophic forgetting} in the online adaptation of depth prediction, a single-task regression problem.
In addition, we compare our online adaptation framework against the state-of-the-art pre-trained depth prediction CNNs to show that our online adapted depth prediction CNN outperforms the depth prediction CNNs that have been trained on a large collection of datasets.
\end{abstract}


\section{Introduction}
\IEEEPARstart{L}{}ong-term adaptation in robotics is challenging yet invaluable for many practical applications, ranging from safe autonomous driving to human-robot interaction.
This paper is concerned with two related but distinct problems: visual SLAM and single- image depth prediction.
Visual SLAM is a problem concerning estimating the structure and camera motion, whereby the camera is attached to a mobile robot or device. 
On the other hand, single-image depth prediction is a method of predicting a depth map from an image. 
In particular, we investigate the problem of visual SLAM with online adaptation by on-demand fine-tuning of a depth prediction convolutional neural network (CNN) and integrating online adapted depth prediction to improve structure and motion estimation.

Given a pre-trained depth prediction CNN---which may not necessarily perform well in a real-world setting---can we still integrate the CNN in a traditional SLAM pipeline such that the depth prediction CNN and the SLAM performance are simultaneously improved?
To address this problem, we propose an on-demand adaptation framework.
That is, we perform SLAM in an environment whose keyframes are used to determine the \textit{quality} of predicted depth and, depending on the \textit{quality}, we switch between fine-tuning of the depth prediction CNN and incorporating fine-tuned depth prediction into global BA to improve SLAM accuracy.
The main contributions of our work are as follows:
\begin{itemize}
	\item We propose a novel online adaptation framework to fine-tune monocular depth prediction on-demand and perform global BA to improve SLAM accuracy. Experimental results in real-world settings demonstrate an improvement in SLAM accuracy after incorporating online adapted depth in global photometric BA.
	\item To perform online adaptation without \textit{catastrophic forgetting}, we investigate the effectiveness of EWC regularization in the online adaptation of a depth prediction CNN, a single-task regression problem, as opposed to heavily studied regularization in multi-task classification problems~\cite{continuallearningsurvey,continuallearningreview}. 
	Experimental results show that use of CNN parameter importance regularization retains the previously learned knowledge better than using only \textit{experience replay}~\cite{robustonlineadapt,comoda,learndepthagainstforgetting}. 
	\item To improve the SLAM accuracy using the adapted depth prediction CNN, we propose using online adapted depth for culling the potential noisy map points before performing global photometric BA, resulting in an improvement in camera tracking and map reconstruction.
\end{itemize}
\IEEEpubidadjcol

\section{Related work}

A modern SLAM algorithm~\cite{pastpresentfutureslam} typically consists of a front-end and a back-end.
Particularly in visual SLAM, the front-end extracts and matches reliable and repeatable features in the images captured by a camera.
Two dominant approaches used in the front-end are indirect~\cite{orbslam3,vitamine} and direct methods~\cite{dso,svo2} whose goal is to define the observation, odometry and optionally loop constraints for the back-end optimization.
Indirect methods preprocess images to generate intermediate feature representation such as handcrafted ORB~\cite{orbdesc} and deep CNN descriptors~\cite{cnndescriptorcomp,selfsupervisedfeatslam,hfnet} to facilitate feature matching.
Alternatively, direct methods operate directly on the raw sensory data---the image pixels---for establishing constraints.
Then, the back-end performs a \textit{maximum a posteriori} (MAP), e.g., BA~\cite{vslamfilter} and pose-graph optimization~\cite{graphslamtutorial}, on the established constraints to obtain optimized SLAM variables consisting of a set of camera poses and the position of 3D points obtained from the matched visual features.

Relevantly, we also review background literature on single-image depth prediction.
Predicting depth from a single image has shown tremendous progress recently with CNNs.
The three main pillars that drive depth prediction accuracy are improved training methods, novel CNN architectures and rich and diverse training datasets.
One of the pioneering works is the supervised learning of global and local depth information with a CNN proposed by Eigen et al., which requires ground truth labels to train the CNN parameters.
Another seminal research direction is unsupervised learning using photometric reconstruction loss from binocular~\cite{gargunsuperviseddepth,monodepth} and motion stereo~\cite{sfmlearner} to overcome the need for ground truth depth.
For improved training accuracy, semi-supervised learning incorporates photometric reconstruction and ground truth depth losses~\cite{semidepth}.
Recently, novel CNN architectures have been proposed to improve depth prediction accuracy, including discretization of the depth output~\cite{dorn,adabins}, 3D packing and unpacking convolutional blocks~\cite{packnet_sfm}, dense prediction transformers~\cite{dpt} and generative adversarial networks~\cite{ganunsuperviseddepth}. 
Furthermore, single-image relative depth prediction, such as MiDaS~\cite{midas} and DiverseDepth~\cite{diversedepth}, has been shown to predict accurate relative depth on diverse scene types by leveraging an extensive collection of training datasets.

To overcome the domain gap of single-image depth prediction, online learning has been used to fine-tune a depth prediction CNN on the images available in the target domain.
These images typically arrive sequentially~\cite{robustonlineadapt,comoda,depthonlineadapt}.
One particular challenge for online fine-tuning is ensuring that learning does not overfit the most \textit{recent} data, a problem known as \textit{catastrophic forgetting}~\cite{lifelonglearning}.
\textit{Experience replay} has been proposed to mitigate \textit{catastrophic forgetting} by inserting randomly sampled past training data into the current training batch~\cite{robustonlineadapt,comoda,learndepthagainstforgetting}.
An alternative solution is to regularize and preserve the CNN parameters that are important to the previously learned tasks~\cite{overcomeforgetting,lifelonglearning}.
In a typical multi-task learning scenario, the importance of the CNN parameters can be measured by the magnitude of the gradients with respect to the loss function or output function through the training on a task~\cite{overcomeforgetting,mas,sicontinuallearn,continuoussitlearning}, which intuitively determines how a perturbation in a CNN parameter affects the loss. 
Instead of measuring the parameter importance after learning a task and regularizing the parameters in learning the next task, Maltoni and Lomonaco~\cite{continuoussitlearning} propose single-incremental-task (SIT) learning, which seeks to estimate and consolidate the parameter importance from batch to batch using synaptic intelligence (SI), so that the previously learned knowledge is retained throughout the learning.
Unlike SIT, which has to deal with learning new instances or classes, we consider single-task learning whose goal is to preserve the previously learned depth from batch to batch by consolidating the parameter importance using a semi-supervised loss, instead of solving classification problems with ground truth labels~\cite{continuoussitlearning}.

Depth prediction by a CNN has been widely used to benefit SLAM algorithms.
A SLAM pipeline can be enhanced by improving feature matching~\cite{cnn-svo} and photometric re-projection accuracies~\cite{dvso} through the use of depth prediction by CNNs.
However, the predicted depth has to be accurate to achieve good performance, an assumption  that is often violated in practical robotics applications.
Therefore, instead of directly incorporating the predicted depth information into front-end tracking and mapping~\cite{cnn-svo,dvso}, we design a VO pipeline that only incorporates online adapted depth information in optimizing the structure and motion estimation.
To this end, we use two common procedures in SLAM optimization: map point culling and global BA.
Removing noisy map points is an essential part of sparse SLAM for preventing erroneous state estimation from being included in optimization (see ORB-SLAM3~\cite{orbslam3} and DSO~\cite{dso});
on the other hand, global photometric BA~\cite{distributedphotoba,photobavslam} has been shown to improve SLAM performance further.
Assuming that online fine-tuning can obtain reasonable depth prediction (without extreme accuracy), we use depth prediction as a hint to cull the potential noisy map points from being included in global photometric BA, resulting in an improved SLAM accuracy.
As for SLAM, we opt for SVO~\cite{svo2} for being a computationally lightweight on an embedded computer, compared to the other state-of-the-art sparse SLAM algorithms~\cite{orbslam3,dso}.

\section{Method}

\begin{figure}[thpb]
	\centering
	\includegraphics[scale=1.0]{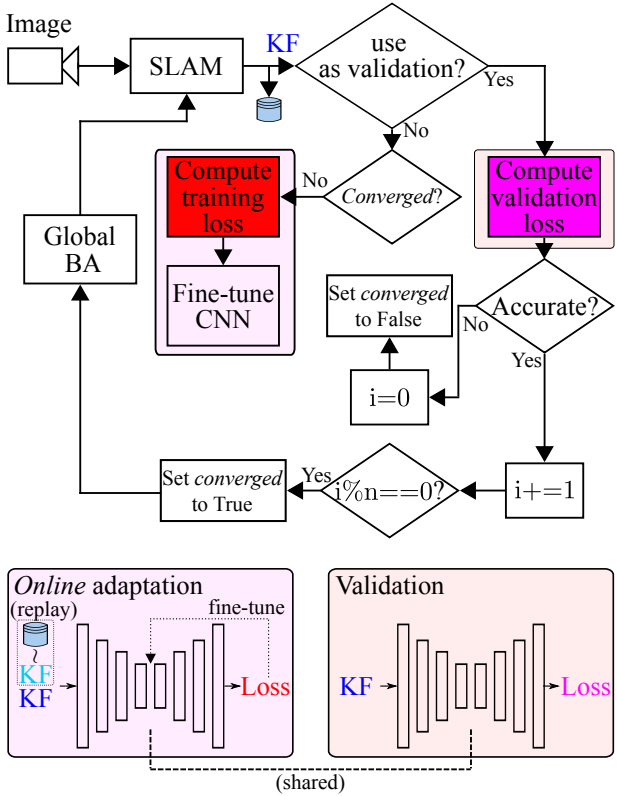}
	\caption{Our proposed online adaptation framework. We use a SLAM algorithm to generate a sequence of keyframes. The keyframes are classified as training or validation to fine-tune a depth prediction CNN and monitor the adaptation progress. If the training is not converged, we use the most recent keyframe and one randomly sampled old keyframe to fine-tune the CNN. Meanwhile, we calculate the validation loss once every $m$ keyframes to determine if the predicted depth maps are accurate and keep track of the number of continuous accurate depth predictions. If the CNN has been accurate for the past $n$ keyframes, we perform global photometric BA. KF: keyframe. }
	\label{fig:pipeline}
\end{figure}

In this section, we outline our proposed online adaptation framework (see Figure~\ref{fig:pipeline}).
The framework consists of SLAM (Section~\ref{subsec:slam}), online CNN depth adaptation (Section~\ref{subsec:cnn_adapt}), and global BA with adapted CNN depth (Section~\ref{subsec:global_ba}).

\subsection{SLAM} \label{subsec:slam}

We use SVO 2.0~\cite{svo2} (its monocular variant with edgelets\footnote{\url{https://github.com/HeYijia/svo_edgelet}}) to generate a set of keyframes from a monocular image stream captured by the camera on a mobile robot.
Associated with each keyframe are an image, a camera pose, and a set of sparse map points.
While the keyframes are being generated, we publish two serialized channels\footnote{The keyframes are converted to ROS messages before they are published. See \url{http://wiki.ros.org/msg} for more information.} over a local network containing the newly created keyframes (image, pose and map points) and keyframe graph (latest optimized camera poses and map points of all keyframes).
The reason for publishing the keyframe graph is that BA, in most cases, improves the camera pose and map point estimations. Therefore, it is beneficial for improving the training loss in the online adaptation.


\subsection{Online CNN depth adaptation}\label{subsec:cnn_adapt}

Our online adaptation is inspired by \textit{early stopping}\footnote{\url{https://www.tensorflow.org/api_docs/python/tf/keras/callbacks/EarlyStopping}}~\cite{earlystopping}, whose goal is to avoid overfitting by stopping the fine-tuning when depth prediction is reasonably accurate, evaluated in terms of validation error (see Algorithm~\ref{alg:adaptation_framework} for detailed steps).
To this end, we validate the depth prediction accuracy once every $m$ incoming keyframes to determine if the fine-tuning has converged.
After determining the stopping criterion, we describe the training datails of the online adaptation algorithm.

\begin{algorithm}
	\caption{online adaptation framework represented in a finite-state machine}\label{alg:adaptation_framework}
	\begin{algorithmic}
		\State \texttt{state} $ = $ \texttt{IDLE}
		\State $t=0$
		\State $n_{\text{converged}} = 0$
		\State $m = 5$ \Comment{Validate every $m$ steps}
		\State $n = 3$ \Comment{\texttt{patience} in the context of \textit{early stopping}}
		\State $\tau_{\text{val}} = 0.2$ \Comment{Validation error threshold}
		\While{not quit}
		\If{ \texttt{state} $ == $ \texttt{IDLE}}
		
		\State \texttt{run\_global\_ba} $ = $ false
		
		\If{$t>0$ and $t\%m == 0$}  
		\State Compute validation error ($\mathcal{L}_{\text{val}}$)
		
		\If{$\mathcal{L}_{\text{val}} < \tau_{\text{val}}$}
		\State $n_{\text{converged}} \mathrel{+}= 1$
		\Else
		\State $n_{\text{converged}} = 0$
		\EndIf
		
		\If{$n_{\text{converged}} > 0$ and $n_{\text{converged}} \% n == 0$} 
		\State \texttt{run\_global\_ba} $ = $ true
		\EndIf
		\Else
		\State \texttt{state} $ = $ \texttt{FINE\_TUNE}
		\EndIf
		\ElsIf{\texttt{state} $ == $ \texttt{FINE\_TUNE}}
		\State Fine-tune the depth prediction CNN
		\State \texttt{state} $ = $ \texttt{IDLE}
		\EndIf
		
		\If{\texttt{run\_global\_ba}}
		\State go through all keyframe and publish the keyframe depth maps to trigger BA in SVO
		\EndIf
		
		\State $t \mathrel{+}= 1$
		\EndWhile
	\end{algorithmic}
\end{algorithm}

To adapt a depth prediction CNN on sequential data without \textit{forgetting}, we improve upon the \textit{experience replay} method (i.e., incorporating \textit{recent} data and randomly selecting \textit{old} data into the training batches~\cite{learndepthagainstforgetting,robustonlineadapt,comoda}) by adding regularization.
Training regularization requires the consideration of the learning of two separate tasks $\mathcal{T}_1$ (learn from \textit{old} data) and $\mathcal{T}_2$ (learn from \textit{recent} data).
The goal is to maximize the training accuracy on both $\mathcal{T}_1$ and $\mathcal{T}_2$, ensuring that we preserve the depth prediction accuracy in the new environment.
To reinforce the previously learned knowledge, we modify EWC~\cite{overcomeforgetting} regularization in the following ways.
First, instead of multi-task learning consolidation, we consolidate the importance of the parameters at each training batch and use it in the next batch, which considers the adaptation progress thus far to be absorbed into the posterior probability, i.e, the importance, of the parameters.
Second, to estimate the posterior probability, we compute the gradient of the negative log-likelihood of the training loss, assuming constant observation noise for the input\footnote{Realistically, observation noise can occur in the violation of brightness constancy assumption and motion blur. This issue can be mitigated by incorporating the heteroscedastic aleatoric uncertainty in training (studied in \cite{d3vo,alexuncertainty}) and modifying a pre-trained depth prediction CNN to predict an additional uncertainty output.} ~\cite{d3vo,alexuncertainty}.
Let $\theta^*$ be the CNN parameters fine-tuned on the last training batch and $\theta$ be the current CNN parameters.
We introduce an additional regularization term to the training loss\footnote{We explain the details in \hyperref[subsec:ewc_append]{Appendix A}. } ~\cite{overcomeforgetting}:
\begin{equation}
\label{eq:ewc_loss}
\mathcal{L}_{\textup{EWC}} = \mathcal{L}_{\textup{train}}  + \sum_i \frac{\beta}{2}\hat{\mathbf{F}}_i (\theta_i - \theta_{i}^*)^2
\end{equation}
with
\begin{align}
\mathbf{F} &= \mathbf{F} + \mathbf{F}^{(j)} \nonumber \\
\hat{\mathbf{F}} &= \text{max}(\frac{\mathbf{F}}{j}, \text{max}_\mathbf{F}), \label{eq:ewc_consolidation}
\end{align}
where $j$ is the number of trained batches and $\mathbf{F}$ the consolidated empirical Fisher information matrix at the end of every training batch.
$\hat{\mathbf{F}}_i$ is importance of the $i$-th parameter, which is obtained from the averaged parameter importance matrix $\hat{\mathbf{F}}$ to ensure the preservation of the previously learned knowledge. 
The empirical Fisher information matrix converges to the Fisher information matrix as more training samples are incorporated~\cite{empiricalfisher}.
$\text{max}_\mathbf{F}$ determines the maximum value of $\hat{\mathbf{F}}$.

\begin{algorithm}
	\caption{Online fine-tuning with parameter importance regularization}
	\label{alg:online_adap}
	\begin{algorithmic}[1]
		\State Init $ \theta$ from a pre-trained depth prediction CNN
		\State $\hat{\mathbf{F}} = 0$ \Comment{parameter importance matrix}
		\While{not done}
		\State $\theta_{\textup{cpy}} = \theta$ \Comment{Make a copy}
		\State Get latest keyframe $\mathcal{K}_i$ from SVO
		
		\State Train the CNN on $\mathcal{K}_i$ and $\mathcal{K}_{j \sim \{0, ..., (i-1)\}}$ with $\mathcal{L}_{\text{EWC}}$ (see Eq.~\ref{eq:ewc_loss}) according to $\theta$ and $\hat{\mathbf{F}}$
		
		\State Consolidate the parameter importance matrix $\hat{\mathbf{F}}$ according to $\theta_{\textup{cpy}}$ and $\theta$ (see Eq.~\ref{eq:ewc_consolidation})
		\EndWhile
		
	\end{algorithmic}
\end{algorithm}

To compute the training loss ($ \mathcal{L}_{\text{train}} $), we use a \textit{semi-supervised} training loss~\cite{monodepth2,semidepth}, which consists of photometric re-projection ($\mathcal{L}_{\textup{photo}}$), sparse depth ($\mathcal{L}_{\textup{sparse\_depth}}$) and smoothness loss ($\mathcal{L}_{\textup{smooth}}$) terms, to be defined below.

%
First, the photometric reprojection error $\mathcal{L}_{\text{photo}}$ is given by:
\begin{align}
\mathcal{L}_{\textup{photo}} &= \frac{1}{\lvert \Omega \rvert} \sum_{\mathbf{p} \in \Omega_s} \argmin_{s} pe(I_i, I_{s \rightarrow i},\mathbf{p}), \text{ and}\\
pe(I_i, I_{s \rightarrow i}, \mathbf{p}) &=  \frac{\alpha}{2} \Big( 1-\textup{SSIM}(I_i, I_{s \rightarrow i})(\mathbf{p}) \Big) \nonumber \\ 
&\qquad + (1 - \alpha) \parallel I_i(\mathbf{p}) - I_{s \rightarrow i}(\mathbf{p}) \parallel_1,
\end{align}
where $pe(\cdot, \cdot, \cdot)$ is the minimum per-pixel photometric re-projection error~\cite{monodepth2} between the reconstructed images from the adjacent keyframes $ s \in \{i-1, i+1\} $ and the $i$-th keyframe.
To reconstruct the target image from an adjacent image $I_{s \rightarrow i}$, we use bilinear sampling at the valid ($\Omega_s$) re-projected pixel locations $\mathbf{p}^\prime$ using depth prediction $D_{i, \textup{CNN}}$, a known camera intrinsics $\mathbf{K}$ and relative keyframe pose transformation $\mathbf{T}_{i \rightarrow s}$ from SVO:
\begin{equation}
\label{eqn:inverse_warp2}
\begin{gathered}
I_{s \rightarrow i}(\mathbf{p}) = I_{s}(\mathbf{p}^{\prime}) \quad \forall \mathbf{p} \in \Omega_s,\\ %
\mathbf{p}^{\prime} \sim \mathbf{K} \mathbf{T}_{i \rightarrow s} D_{i, \textup{CNN}}(\mathbf{p}) \mathbf{K}^{-1} \mathbf{p}.
\end{gathered}
\end{equation}
%

Then, to define $\mathcal{L}_{\text{sparse\_depth}}$, we project the sparse map points generated by SVO in each keyframe to form sparse depth maps as \textit{ground truth} labels~\cite{semidepth,pseudorgbd}:
\begin{equation}
\mathcal{L}_{\textup{sparse\_depth}} = \frac{1}{\lvert \Omega_{i, \textup{sparse}} \rvert} \sum_{\mathbf{p} \in \Omega_{i, \textup{sparse}}} \lvert \frac{1}{D_{i, \textup{CNN}}(\mathbf{p})} - \frac{1}{D_{i, \textup{sparse}}(\mathbf{p})} \rvert,
\end{equation}
where $D_{i, \textup{sparse}}$ is the sparse depth map of keyframe $i$, $\Omega_{i, \textup{sparse}}$ a set of re-projected (sub)pixel locations containing depth values of their corresponding map points from SVO, and $\lvert \Omega_{i, \textup{sparse}} \rvert$ the number of valid re-projections. 
By using inverse depth, near depth is penalized more heavily than far depth.

%
Finally, we compute the edge-aware smoothness loss $\mathcal{L}_{\text{smooth}}$ using the image gradient:
\begin{align}
\mathcal{L}_{\textup{smooth}} &= \frac{1}{N} \sum_{\mathbf{p}}  \lvert \partial_x D_{i, \textup{CNN}}(\mathbf{p}) \rvert e^{- \lvert \partial_x I_i(\mathbf{p}) \rvert} \nonumber \\
&\qquad + \lvert \partial_y D_{i, \textup{CNN}}(\mathbf{p}) \rvert e^{- \lvert \partial_y I_i(\mathbf{p}) \rvert}
\end{align}
where $\partial$ is the gradient operator. 

Combining the three loss terms, the final training loss is
\begin{equation}
\mathcal{L}_{\textup{train}} = \mathcal{L}_{\textup{photo}} + \lambda_1 \mathcal{L}_{\textup{sparse\_depth}} + \lambda_2 \mathcal{L}_{\textup{smooth}}
\end{equation}
and averaged across the most recent keyframe and a randomly selected old keyframe (\textit{experience replay}).
$\lambda_1$ and $\lambda_2$ are the weighting parameters.
Algorithm~\ref{alg:online_adap} details the steps involved in the online fine-tuning.
To validate the training accuracy, we use the sparse depth loss term:
\begin{equation}
\label{eq:val_err}
\mathcal{L}_\textup{val} = \mathcal{L}_\textup{sparse\_depth}.
\end{equation}

\subsection{Global BA} \label{subsec:global_ba}

\begin{figure}
	\centering
	\includegraphics[scale=0.8]{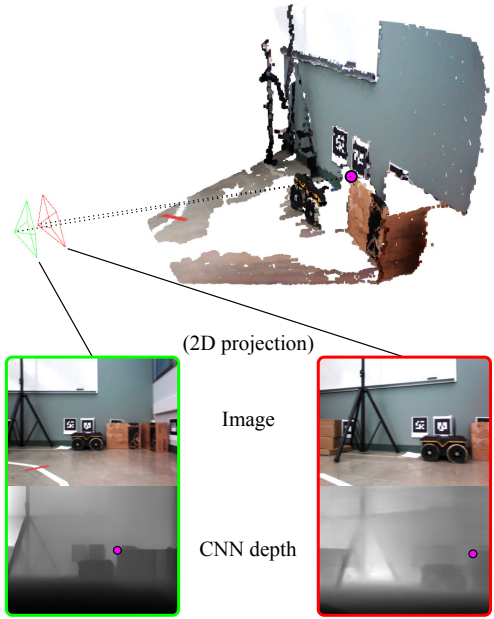}
	\caption{Assuming the magenta map point is observed in two keyframes (the red and green camera frustums), a host keyframe is selected based on the validation loss ($\mathcal{L}_{\text{val}}$) of the predicted CNN depth, and in this case, the green keyframe has a lower $\mathcal{L}_{\text{val}}$ and hence is being selected as the host keyframe of the magenta map point. }
	\label{fig:mp_host_correctness}
\end{figure}

To optimize the structure and motion, we employ the traditional photometric BA~\cite{photobavslam}, which jointly optimize the keyframe poses and map points.
However, we found that the map points generated by SVO, albeit the sparsest amongst the state-of-the-art visual SLAM algorithms~\cite{orbslam3,dso}, still contain noisy and redundant map points.
Therefore, with the learned depth information, we introduce a map point culling step before performing global photometric BA.
To determine if a map point should be culled, we identify a host keyframe for the map point (see Figure~\ref{fig:mp_host_correctness}) and check if the depth of the map point is within the \textit{correctness} range:
\begin{equation}
g(d_{\text{mp}}, d_{\text{CNN}}) =
\begin{cases}
1,  & \text{if } | d_{\text{mp}} - d_{\text{CNN}} | < \gamma d_{\text{CNN}} \enspace \text{or} \\
& \quad d_{\text{CNN}} > d_{\text{max}} \\
0,  & \text{otherwise}
\end{cases}
\end{equation}
where $d_{\text{CNN}}$ and $d_{\text{mp}}$ are the predicted CNN depth and the depth of the map point in the host keyframe, $d_{\text{max}}$ a threshold that defines an \textit{effective} depth range of the CNN depth values (similar to a depth sensor) to avoid far map points from being culled prematurely.
$g(\cdot)$ evaluates if the map point is valid.
Then, we use the standard photometric BA formulation~\cite{photobavslam} to globally optimize the structure and camera poses (collectively defined as the SLAM variables $\mathcal{X}$) given by:
\begin{equation}
\mathcal{X}^* = \argminA_{\mathcal{X}} \frac{1}{2} \sum_{k}  \| \mathbf{e}_{k} \|^2_2
\end{equation}
with
\begin{equation}
\mathbf{e}_k = \mathbf{z}_k - h_k(\mathcal{X}_k)
\end{equation}
where $\mathbf{e}_k$ is the photometric error induced by a subset of the SLAM variables $\mathcal{X}_k \subseteq \mathcal{X}$, $\mathbf{z}_k$ the reference image patch and $h_k(\cdot)$ the measurement model to obtain the re-projected image patch.

\section{Evaluation}

\begin{figure}
	\centering
	\includegraphics[scale=1.0]{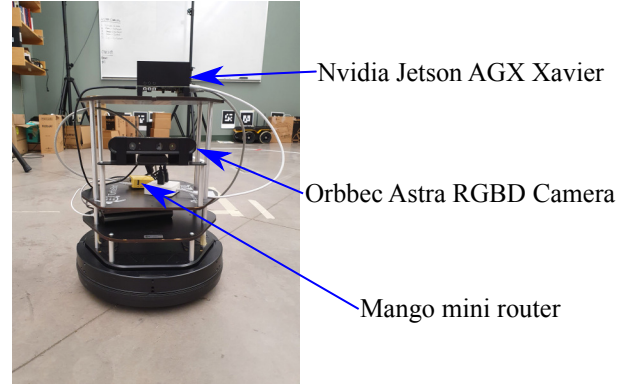}
	\caption{A TurtleBot equipped with an Nvidia Jetson AGX Xavier and an Orbbec Astra RGBD camera. A Mango mini router is used to create a local wireless network to communicate between the Jetson and a laptop. }
	\label{fig:robot_setup}
\end{figure}

To evaluate the performance of our proposed method for online CNN adaptation experimentally, our mobile robot hardware setup consists of the following main components: a TurtleBot, an Nvidia Jetson AGX Xavier, an Orbbec Astra RGBD camera, and a laptop\footnote{Specifications: Intel 7820HK CPU and Nvidia GTX 1070 GPU}, as shown in Figure~\ref{fig:robot_setup}.
With the Jetson and laptop connected to the same wireless network, we run the SLAM process (tracking, mapping, and BA) on the Jetson and the fine-tuning process on the laptop. 

To illustrate the effectiveness of the online adaptation, we use Monodepth2's \texttt{mono+stereo\_640x192} pre-trained CNN model, which has been trained on outdoor scenes fine-tuned in an indoor environment.
For the network to learn, we use Adam optimizer with a learning rate of $10^{-3}$, and set the weighting of the loss function $\lambda_1$, $\lambda_2$, $\alpha$ and $\beta$ at $0.1$, $0.1$, $0.85$ and $5 \times 10^7$, respectively. For EWC consolidation of parameter importance matrix, $\text{max}_F $ is set to $0.001$.

In SVO, the main change we made is the increased number of tracked features in the keyframes.
We have modified the following parameter values: $\texttt{max\_fts}=500$, $\texttt{grid\_size}=20$, and $\texttt{core\_kfs}=5$.
For performing map point culling, we define the \textit{correctness} range by setting $\gamma = 0.5$ and $d_{\text{max}} = 1.5$. 
This particular design allows for more map points to be generated for fine-tuning the depth prediction CNN.
The online adapted depth is then used for removing the potential noisy map points.
After the noisy point removal, each map point has a maximum of five photometric re-projections to the nearby keyframes, including the existing 3D-2D constraints (i.e., the list of observed keyframes for the map point), for performing global photometric BA in a separate thread.
For each re-projection edge, we get a $3 \times 3$ image patch around the re-projected image coordinates to compute the photometric errors.


In the following, we present experimental results to validate the performance of our proposed method.
Section~\ref{subsec:mini_dataset} details the dataset that we collect in our laboratory and its purposes.
Section~\ref{subsec:online_ada} compares and contrasts our proposed online adaptation method with regularization against the state-of-the-art methods.
Section~\ref{subsec:reg_ada} conducts an ablation study to compare different online adaptation schemes in overcoming \textit{catastrophic forgetting}. 
Section~\ref{subsec:recon_accu} presents quantitative and qualitative results of our proposed map point culling with online adapted depth and global photometric BA to improve SLAM accuracy.
Section~\ref{subsec:online_vs_pretrained} analyzes the accuracy of online adapted and pre-trained depth prediction to show the advantage of using online adaptation to improve SLAM performance.
And lastly, Section~\ref{subsec:timing_eval} evaluates the runtimes of our proposed online adaptation framework.


\subsection{Laboratory dataset}\label{subsec:mini_dataset}

The reason for collecting our own dataset is that the image sequences in existing benchmarking datasets~\cite{tumrgbddataset,nyuv2dataset,scannetdataset} are not long enough to evaluate and illustrate the effectiveness of our proposed online adaptation. 
Our dataset contains two sequences (dubbed \texttt{Lab1} and \texttt{Lab2}) that we collected in our laboratory (see Figure~\ref{fig:mp_culling_qualitative} for the images of the captured environment).
We record both sequences in the same environment, with the difference being the \texttt{Lab2} sequence (16366 images) is longer than the \texttt{Lab1} sequence (10611 images).

\subsection{Online adaptation}\label{subsec:online_ada}

To evaluate the effectiveness of our proposed online adaptation, we compare our method against the state-of-the-art methods: a SLAM-based approach by Luo et al.~\cite{depthonlineadapt} (Section~\ref{subsubsec:slam_ada}) and a learning-based approach by Kuznietsov et al.~\cite{comoda} (Section~\ref{subsubsec:ete_ada}).
For the comparison, we adopt the datasets and performance metric in \cite{depthonlineadapt}.
The datasets are the ICL-NUIM~\cite{iclnuimdataset} and TUM RGB-D~\cite{tumrgbddataset} datasets and the performance metric is the percentage of overall depth pixels below $10\%$ relative errors, i.e., $\frac{\lvert D - D_{gt} \rvert}{D_{gt}} < 0.1$.

\subsubsection{SLAM-based online adaptation}\label{subsubsec:slam_ada}

Table~\ref{table:ada_slam_accuracy} compares the overall depth accuracy between our method and a similar method by Luo et al.~\cite{depthonlineadapt}.
Our method improves the overall depth accuracy by around $23\%$ (the last two columns), compared to $4\%$ (2nd and 3rd column) by Luo et al.~\cite{depthonlineadapt}.
The performance gap could be due to the number of keyframes used in the fine-tuning;
we use all the keyframes generated by SVO~\cite{svo2}, whereas Luo et al.~\cite{depthonlineadapt} use the keyframe pairs by LSD-SLAM~\cite{lsdslam} that have a dominant horizontal motion as simulated static stereo pairs.
Besides, Luo et al.~\cite{depthonlineadapt} perform online adaptation using most \textit{recent} keyframes only, which performs the worst in the online adaptation scheme comparison (see Section~\ref{subsec:reg_ada}).
Note that Luo et al.~\cite{depthonlineadapt} evaluate the keyframe depth accuracy based on the fine-tuned CNN models in different fine-tuning stages, which may result in different overall depth correctness should the final fine-tuned CNN model be used in the evaluation.

\begin{table}
	\begin{center}
		\caption{A comparison between the overall depth accuracy of our method and Luo et al.'s~\cite{depthonlineadapt} SLAM-based online adaptation on the ICL-NUM~\cite{iclnuimdataset} and TUM RGB-D~\cite{tumrgbddataset} datasets. (TUM/seq1: fr3\_long\_office\_household, TUM/seq2: fr3\_nostructure\_texture\_near\_withloop, TUM/seq3: fr3\_structure\_texture\_far)}
		\label{table:ada_slam_accuracy}
		\begin{tabular}{c | c c | c c} 
			\Xhline{2\arrayrulewidth}
			& \multicolumn{4}{c }{Percentage of correct depth (\%)} \\
			\cline{2-5}
			Sequence         & Pre-trained~\cite{depthonlineadapt}      & \cite{depthonlineadapt} & Pre-trained$^*$  & Ours$^*$  \\ 
			\hline
			ICL/office0      & 19.117            & \textbf{22.206}            & 8.766   & 12.541           \\ 
			ICL/office1      & 28.086            & \textbf{31.289}            & 19.739  & 21.870    \\  
			ICL/office2      & 21.695            & 21.695                     & 4.591   & \textbf{42.244}          \\
			ICL/living0      & 18.680            & 23.278                     & 7.726   & \textbf{41.375 }          \\
			ICL/living1      & 21.071            & 22.774                     & 8.518   & \textbf{49.075}      \\ 
			ICL/living2      & 16.150            & 20.995                     & 13.509  & \textbf{25.159}            \\
			TUM/seq1         & 18.208            & 20.259                     & 10.999  & \textbf{23.861}    \\
			TUM/seq2         & 25.796            & 29.014                     & 12.678  & \textbf{52.162}    \\
			TUM/seq3         & 20.668            & 30.156                     & 10.295 & \textbf{37.848 }     \\ 
			\hline
			Average          & 21.052            & 24.630                  & 10.758  & \textbf{34.015}     \\
			\Xhline{2\arrayrulewidth}
			\multicolumn{5}{l}{$^*$ Scaled to ground truth depth}
		\end{tabular}
	\end{center}
\end{table}

\subsubsection{Learning-based online adaptation}\label{subsubsec:ete_ada}

Table~\ref{table:ete_accuracy} compares our method against a learning-based online adaptation method, CoMoDa\footnote{We generate results using CoMoDa's~\cite{comoda} open-source code\footnote{https://github.com/Yevkuzn/CoMoDA}}~\cite{comoda}.
Overall, our proposed SLAM-based method outperforms CoMoDa by around $5\%$ (the last two columns).
One particular challenge in end-to-end online adaptation is the simultaneous fine-tuning of depth and pose prediction CNNs, and the pose prediction CNN is not trained in the tested indoor environments;
on the contrary, we obtain camera poses from SVO~\cite{svo2}.
Furthermore, accurate camera pose estimation is essential when performing online adaptation (see Equation~\ref{eqn:inverse_warp2}).
Plus, CoMoDa does not use regularization in online adaptation, which is a reason that it performs worse than our method (see Section~\ref{subsec:reg_ada}).

\begin{table}
	\begin{center}
		\caption{A comparison between the overall depth accuracy of our method and CoMoDa's~\cite{comoda} end-to-end online adaptation on the ICL-NUM~\cite{iclnuimdataset} and TUM RGB-D~\cite{tumrgbddataset} datasets. Both our method and CoMoDa~\cite{comoda} are fine-tuned on the same pre-trained CNN model, \texttt{mono+stereo\_640x192}~\cite{monodepth2}. (TUM/seq1: fr3\_long\_office\_household, TUM/seq2: fr3\_nostructure\_texture\_near\_withloop, TUM/seq3: fr3\_structure\_texture\_far)}
		\label{table:ete_accuracy}
		\begin{tabular}{c | c c c} 
			\Xhline{2\arrayrulewidth}
			& \multicolumn{3}{c }{Percentage of correct depth (\%)} \\
			\cline{2-4}
			Sequence         & Pre-trained$^*$  & Ours$^*$   & CoMoDa$^*$~\cite{comoda}  \\ 
			\hline
			ICL/office0      &   13.660 &   \textbf{32.98}   &   21.054        \\ 
			ICL/office1      &   22.453 &   38.383  &   \textbf{41.623} \\  
			ICL/office2      &  22.514  &   \textbf{44.790}  &   37.374          \\
			ICL/living0      &  17.659  &   \textbf{48.158}  &   35.164           \\
			ICL/living1      &  22.585  &   \textbf{51.531}  &   46.148           \\ 
			ICL/living2      &  21.606  &   \textbf{35.757}  &   35.571            \\
			TUM/seq1         &  16.931  &   27.915  &   \textbf{33.120}        \\
			TUM/seq2         &  17.722  &   \textbf{41.782}  &   33.708       \\
			TUM/seq3         &  23.026  &   \textbf{50.339}  &   39.769               \\ 
			\hline
			Average          &  19.795  &   \textbf{41.293}  &   35.948            \\
			\Xhline{2\arrayrulewidth}
			\multicolumn{4}{l}{$^*$Median scaling to ground truth depth}
		\end{tabular}
	\end{center}
\end{table}

\subsection{Learning against \textit{catastrophic forgetting} in online adaptation}\label{subsec:reg_ada}

To evaluate the effectiveness of our proposed regularization in online adaptation, we perform an ablation study to measure the impact of different adaptation schemes on alleviating \textit{catastrophic forgetting}.
To this end, we examine the overall depth accuracy (using the same $10\%$ relative error as the metric) on the ICL-NUIM~\cite{iclnuimdataset} and TUM RGB-D~\cite{tumrgbddataset} datasets.

\begin{figure} 
	\centering
	\includegraphics[scale=1.0]{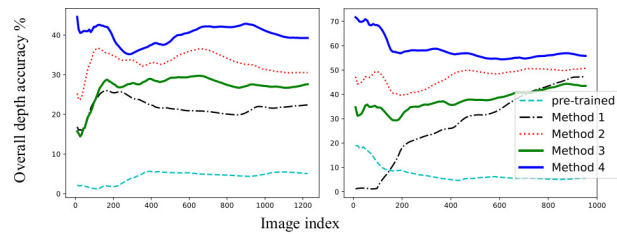}
	\caption{A comparison of different online adaptation schemes tested on the ICL-NUIM~\cite{iclnuimdataset} \texttt{of\_kt3} (left) and \texttt{lr\_kt1} (right) using the final online adapted CNN. Adaptation accuracy is measured by the averaged percentage of overall depth accuracy over all frames up to the frame. Method 1: fine-tuning on most \textit{recent} keyframes only; Method 2: fine-tuning on the most \textit{recent} keyframe with \textit{experience replay}; Method 3: fine-tuning on the most \textit{recent} keyframe with regularization; Method 4: fine-tuning on the most \textit{recent} keyframe with \textit{experience replay} and regularization. }
	\label{fig:ablation_err}
\end{figure}

Table~\ref{table:ada_schemes} shows that performing online adaptation using only most \textit{recent} keyframes has the worst overall depth accuracy (3rd column), compared to using the most \textit{recent} keyframe with \textit{experience replay} and regularization (the last two columns).
Moreover, Figure~\ref{fig:ablation_err} illustrates \textit{catastrophic forgetting} on the \texttt{lr\_kt1} sequence, where the CNN is overfitted to the most \textit{recent} keyframes in the sequence.
In general, a combination of \textit{experience replay} and regularization in online adaptation has the best overall depth accuracy (see the last column in Table~\ref{table:ada_schemes} and Figure~\ref{fig:ablation_err}).
Thus, it validates the effectiveness of our proposed online adaptation scheme.
However, for a short sequence, regularization can inhibit changes in the CNN parameters and hurt the online adaptation accuracy (2nd last row in Table~\ref{table:ada_schemes}).
For the sake of completeness, we compare the online adaptation performances using the three popular regularization techniques\footnote{See \hyperref[subsec:si_mas_reg]{Appendix B} for more information. The regularization strength ($\lambda$ in Eq.~\ref{eq:si_mas_reg}) for SI and MAS are set to 1.0 and 0.5, respectively.}: EWC~\cite{overcomeforgetting}, MAS~\cite{mas} and  SI~\cite{sicontinuallearn}, and they perform similarly.

\begin{table}
	\begin{center}
		\caption{An ablation study on different online adaptation schemes using our proposed framework on the ICL-NUM~\cite{iclnuimdataset} and TUM RGB-D~\cite{tumrgbddataset} datasets. (TUM/seq1: fr3\_long\_office\_household, TUM/seq2: fr3\_nostructure\_texture\_near\_withloop, TUM/seq3: fr3\_structure\_texture\_far)}
		\label{table:ada_schemes}
		\begin{tabular}{c | c c c c } 
			\Xhline{2\arrayrulewidth}
			& \multicolumn{4}{c }{Percentage of correct depth (\%)} \\
			\cline{2-5}
			Sequence         & Pre-trained & Ours(a)   &   Ours(b) & Ours(c)  \\ 
			\hline
			ICL/office0      & 8.766   & 7.249  &   \textbf{12.734}       & 12.541         \\ 
			ICL/office1      & 19.739  & 4.034   & \textbf{22.613}       & 21.870 \\  
			ICL/office2     & 4.591   & 21.449  & 34.609        & \textbf{42.244}          \\
			ICL/living0      & 7.726   & 13.179  & 22.883        & \textbf{41.375 }         \\
			ICL/living1      & 8.518   & 38.246  & 41.149        & \textbf{49.075}           \\ 
			ICL/living2      & 13.509  & 9.110  & 13.117       & \textbf{25.159}        \\
			TUM/seq1         & 10.999  & 16.012  & 19.782       & \textbf{23.861}        \\
			TUM/seq2         & 12.678  & 14.457  & 37.539       & \textbf{52.162}      \\
			TUM/seq3         & 10.295  & 49.515  & \textbf{51.545}       & 37.848             \\ 
			\hline
			Average          & 10.758  & 19.250  & 28.441       & \textbf{34.015}        \\
			\Xhline{2\arrayrulewidth}
			\multicolumn{5}{l}{(a) Online adaptation with two most \textit{recent} keyframes}  \\
			\multicolumn{5}{l}{(b) Online adaptation with the most \textit{recent} keyframe and }  \\
			\multicolumn{5}{l}{\qquad \textit{experience replay}}  \\
			\multicolumn{5}{l}{(c) Online adaptation with the most \textit{recent} keyframe,  } \\
			\multicolumn{5}{l}{\qquad \textit{experience replay} and regularization}  \\
		\end{tabular}
	\end{center}
\end{table}

\subsection{Effect of global photometric BA with map point culling on SLAM accuracy}\label{subsec:recon_accu}

After the depth prediction CNN has been fine-tuned to a reasonable accuracy, our following proposed method incorporates the online adapted depth into removing potential noisy map points and then performs photometric global BA to improve SLAM performance.

\subsubsection{Camera tracking error}
To measure the camera tracking performance, we use the standard absolute trajectory RMSE (ATE) as the performance metric on the ICL-NUIM~\cite{iclnuimdataset} and TUM RGB-D~\cite{tumrgbddataset} datasets, which contains ground truth camera poses for the evaluation.
Table~\ref{table:ate_ba} compares the ATEs obtained from SVO and SVO with our proposed map point culling and global photometric BA.
In general, performing global photometric BA leads to a lower overall ATE than that of without the global BA (compare the last two columns in Table~\ref{table:ate_ba}).
Our proposed photometric global BA with map point culling reduces the ATE by more than $50\%$ (from 0.029 to 0.011) on the \texttt{TUM/seq2} sequence, likely due to the better online adaptation of depth prediction (check the corresponding rows in Table~\ref{table:ada_schemes}).

\begin{table}
	\begin{center}
		\caption{A comparison of the camera tracking ATEs with and without global photometric BA on the ICL-NUM~\cite{iclnuimdataset} and TUM RGB-D~\cite{tumrgbddataset} datasets. (TUM/seq1: fr3\_long\_office\_household, TUM/seq2: fr3\_nostructure\_texture\_near\_withloop, TUM/seq3: fr3\_structure\_texture\_far)}
		\label{table:ate_ba}
		\begin{tabular}{c | c c  } 
			\Xhline{2\arrayrulewidth}
			Sequence         & Without BA$^*$ & With BA$^*$   \\ 
			\hline
			ICL/office0      & 0.363   & \textbf{0.323}        \\ 
			ICL/office1      & \textbf{0.196}  & 0.204  \\  
			ICL/office2     & \textbf{0.170}   & 0.171         \\
			ICL/living0      & \textbf{0.220}   & 0.227       \\
			ICL/living1      & 0.032   & \textbf{0.026}       \\ 
			ICL/living2      & 0.161  & \textbf{0.158}     \\
			TUM/seq1         & 0.088 & \textbf{0.073}    \\
			TUM/seq2         & 0.029  & \textbf{0.011}   \\
			TUM/seq3         & \textbf{0.011}  & 0.012     \\ 
			\hline
			Average          & 0.141  & \textbf{0.134}      \\
			\Xhline{2\arrayrulewidth}
			\multicolumn{3}{l}{$^*$Aligned to ground truth trajectory}
		\end{tabular}
	\end{center}
\end{table}

\subsubsection{Map reconstruction error} \label{subsubsec:map_err}

To highlight the effectiveness of the online adaptation, we evaluate the accuracy of the SVO map on long sequences, \texttt{Lab1} and \texttt{Lab2}, from our Laboratory dataset.
We use the scale-invariant inverse depth error as the performance metric~\cite{demon} given by:
\begin{equation}
e_{\text{si}} = \sqrt{\frac{1}{N} \sum_i d_i^2 - \frac{1}{N^2}(\sum_i d_i)^2}
\end{equation}
where $d_i = \log z_i - \log z_i^*$, and the superscript $^*$ indicates the ground truth depth.

\begin{figure} 
	\centering
	\includegraphics[scale=0.9]{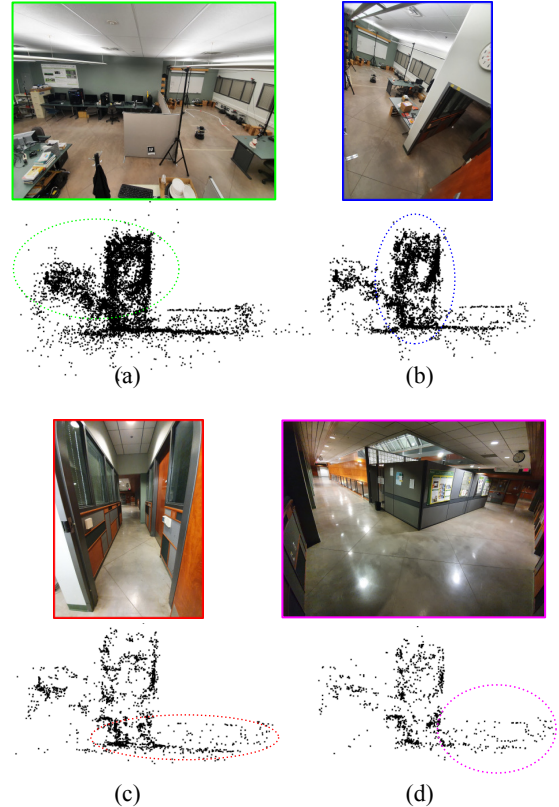}
	\caption{Qualitative comparisons between different \textit{correctness} thresholds used in map point (MP) culling: (a) no culling, (b) MP culling with $\alpha = 0.5$, (c) MP culling with $ \alpha = 0.25 $ and (d) MP culling with $ \alpha = 0.15 $. }
	\label{fig:mp_culling_qualitative}
\end{figure}

To allow for a more accurate online adaptation accuracy, we increase the number of map points by tweaking the \texttt{max\_fts} parameter in SVO.
Table~\ref{table:mps_density_err} shows that the SVO map accuracy is improved by our proposed map point culling with online adapted depth prediction and global photometric BA (compare the $e_{\text{si}}$ with and without BA in Table~\ref{table:mps_density_err} and Figure~\ref{fig:mp_culling_qualitative}(a) and \ref{fig:mp_culling_qualitative}(b)).
The improved map accuracy also comes at the expense of removing good map points, which can be mitigated by having a training scheme with better online adaptation performance.
As a result, our proposed online adaptation framework achieves a mutual adaptation of depth prediction and visual SLAM.

\begin{table}
	\begin{center}
		\caption{A comparison of the number of points and depth error of the SVO map with and without our proposed map point culling and global photometric BA. A larger \texttt{max\_fts} generates more map points in SVO.}
		\label{table:mps_density_err}
		\resizebox{\columnwidth}{!}{%
			\begin{tabular}{ c | c | c | c | c | c | c} 
				\Xhline{2\arrayrulewidth}
				& MP  & SVO param & \multicolumn{2}{c |}{ \texttt{Lab1} }&  \multicolumn{2}{c}{ \texttt{Lab2} }     \\ %
				\cline{4-7}
				BA & culling & (\texttt{max\_fts})  & No. map points & $e_{\text{si}}$ &  No. map points & $e_{\text{si}}$   \\
				\hline
				& & 120 & 13992 & 0.533 & 19856 & 0.607 \\
				\checkmark & & 120 & 14469 & 0.545 & 20464 & 0.625 \\
				\checkmark & \checkmark & 120 & 7435  & 0.473 & 9226 & 0.495 \\
				\hline
				& & 500 & 19179 & 0.527 & 25743 & 0.540 \\
				\checkmark & & 500 & 19208 & 0.515 & 23998 & 0.528 \\
				\checkmark & \checkmark & 500 & 9916  & 0.493 & 12686 & 0.472 \\
				\Xhline{2\arrayrulewidth}
				\multicolumn{5}{l}{Note: \texttt{max\_fts} is set at 120 by default in SVO. $$} \\
				\end{tabular}
			}
	\end{center}
\end{table}



\subsection{Online adaptation vs. relative depth prediction}\label{subsec:online_vs_pretrained}

A competing idea to our online adaption for resolving the domain gap in single-image depth prediction is to train a network with a large number of datasets across multiple domains to improve its generalization.
Unlike absolute depth prediction, which is trained in a narrow domain, state-of-the-art relative depth prediction CNN models, such as MiDaS~\cite{midas} and DiverseDepth~\cite{diversedepth}, have been trained on an extensive collection of datasets.
Therefore, to compare between online adapted depth and pre-trained depth prediction, we use the same scale-invariant inverse depth error ($e_{\text{si}}$) described in Section~\ref{subsubsec:map_err}.

\begin{figure*}
	\centering
	\includegraphics[scale=1.0]{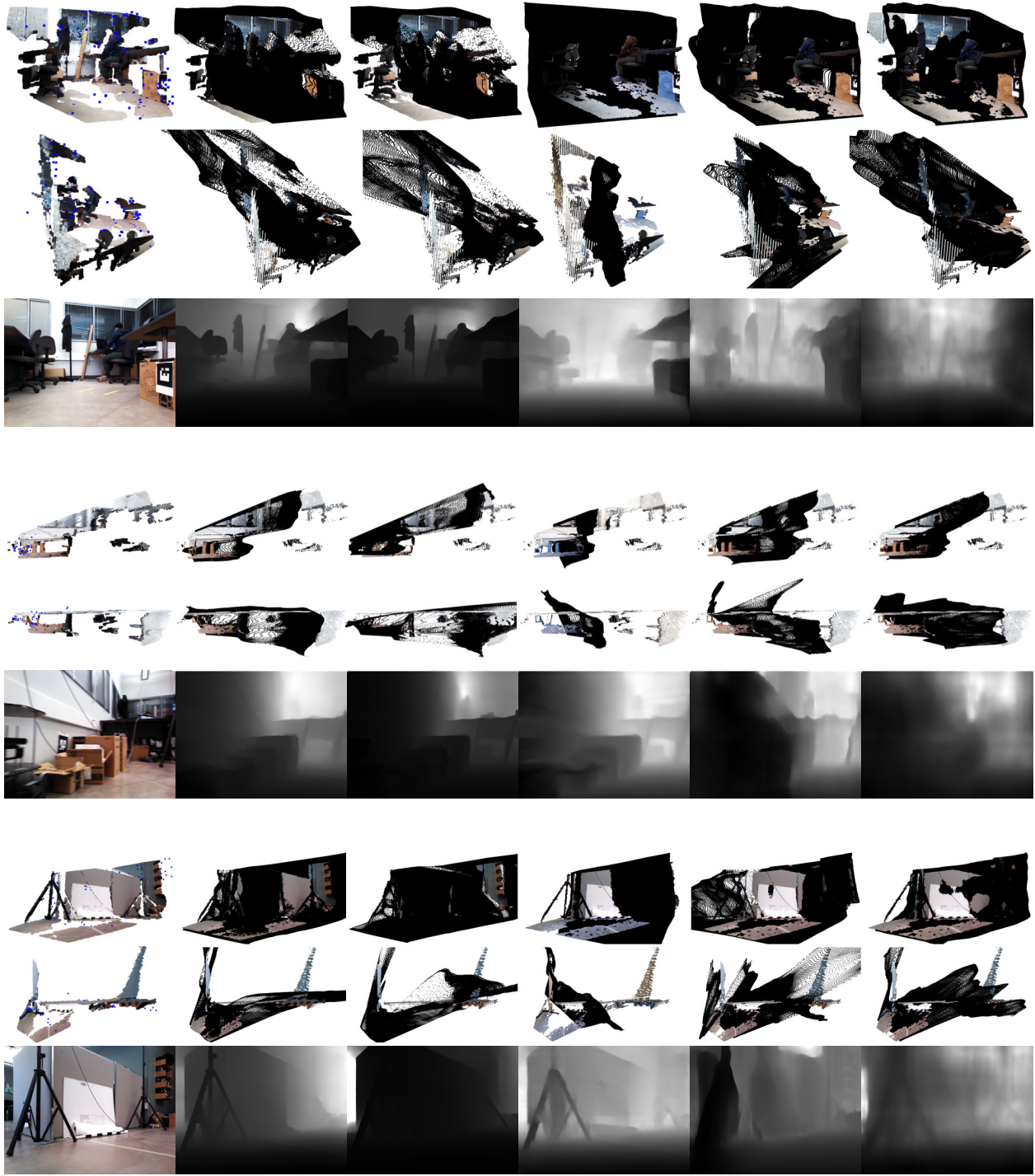}
	\caption{A qualitative comparison of the back-projected point clouds (shown in black) between (from left to right) ground truth depth with SVO map points (in blue), MiDaS \texttt{v2.1}, MiDaS \texttt{v3.0}, DiverseDepth, pre-trained Monodepth2, and our online adapted Monodepth2. From top to bottom: first and second viewpoint  of the back-projected depth maps and the predicted depth maps by the aforementioned CNN models. The predicted depth maps are scaled to ground truth. Best viewed digitally. }
	\label{fig:ablation_depth_accuracy}
\end{figure*}

Table~\ref{table:scale_invariant_err} shows a quantitative comparison of the depth errors of MiDaS, DiverseDepth and our fine-tuned Monodepth2 CNN model.
On both sequences (2nd and 3rd column), our method has the lowest depth error in comparison with the relative depth prediction CNNs\footnote{For relative depth prediction to achieve maximum accuracy, accurate scale- and shift-correction for each predicted depth map is required~\cite{deeprelativefusion,diversedepth}} (MiDaS and DiverseDepth) (see Figure~\ref{fig:ablation_depth_accuracy} for a qualitative comparison).
Due to the similarity between \texttt{Lab1} and \texttt{Lab2}, the depth prediction errors of MiDaS, DiverseDepth and pre-trained Monodepth2 are similar in both sequences.
Conversely, our method can further boost the depth prediction accuracy (see the last row of Table~\ref{table:scale_invariant_err}) resulting from our proposed online adaptation, especially on the longer sequence (\texttt{Lab2}).



\begin{table}
	\begin{center}
		\caption{A comparison of depth errors of MiDaS~\cite{midas}, DiverseDepth~\cite{diversedepth} and our online adapted Monodepth2~\cite{monodepth2} CNN model. }
		\label{table:scale_invariant_err}
		\begin{tabular}{c | c | c } 
			\Xhline{2\arrayrulewidth}
			& \multicolumn{2}{c}{$e_{\text{si}}$ }     \\ %
			\cline{2-3}
			Method  &   \texttt{Lab1}    & \texttt{Lab2} \\
			\hline
			MiDaS (\texttt{V2.1})    & 0.271    & 0.256  \\  
			MiDaS (\texttt{V3.0})    & 0.315    & 0.322  \\
			DiverseDepth             & 0.450    & 0.435 \\
			\hline
			Pre-trained              & 0.401    & 0.399 \\
			Ours                     & \textbf{0.235}    & \textbf{0.219} \\
			\Xhline{2\arrayrulewidth}
			\multicolumn{3}{l}{$^\dagger$ only the re-projected (sub)pixels} \\
		\end{tabular}
	\end{center}
\end{table}

\subsection{Runtime evaluation} \label{subsec:timing_eval}
The two main time-consuming processes in our proposed online adaptation framework are depth prediction and fine-tuning.
On average, depth prediction and one fine-tuning iteration take about 18 ms and 250 ms to process, respectively.
Incidentally, the global photometric BA time consumption varies according to the total number of photometric re-projections between the map points and keyframes, but real-time processing should not be affected as the BA process itself runs on another thread.

\section{Conclusion}
In this paper, we have addressed a practical robotics problem concerning the use of single-image depth prediction to improve SLAM performance.
Particularly, we have proposed a novel online adaptation framework in which the fine-tuning is enhanced with regularization for retaining the previously learned knowledge while the CNN is being trained continually.
Furthermore, we have investigated the use of regularization in the learning of depth prediction, a single-task regression problem, with quantitative results and an ablation study to show its effectiveness against \textit{catastrophic forgetting}.
Also, we have demonstrated the use of fine-tuned depth prediction for map point culling before running global photometric BA, resulting in improved camera tracking and map reconstruction.
Lastly, we have compared our online adaptation framework against state-of-the-art depth prediction CNNs that have been trained on a large number of datasets across domains, showing that our online adapted depth prediction CNN has a lower depth error, especially after performing online adaptation on a long sequence.


One main shortcoming is that SVO is a visual odometry system and is prone to drift over time.
For long-term adaptation problems, it would be better to incorporate loop closure and establish feature correspondences among the closed-loop keyframes to enhance the map reconstruction accuracy while the CNN is being online adapted.
In return, the long-term adapted depth prediction CNN can be used to improve the SLAM accuracy further.


\section*{APPENDIX} \label{sec:appendix}

\subsection{Adapted EWC regularization for single-task regression problem} \label{subsec:ewc_append}

To perform online adaptation incrementally on a single task, we measure and consolidate the importance of the CNN parameters at each iteration, allowing for the preservation of learned knowledge.
The EWC loss comprises the training loss and a regularizer to fine-tune the CNN parameters while preserving the important parameters measured through consolidation.


Formally, let $E$ be the loss function of the problem at hand and assuming $E$ to be locally smooth. we may expand the function about $\theta_0$ using Taylor series approximation:
\begin{align}
E(\theta) &= E(\theta_0) + \mathbf{J}_\theta(\theta_0) (\theta - \theta_0)  \nonumber \\
&\qquad + \frac{1}{2!} (\theta - \theta_0)^T \mathbf{H}_\theta(\theta_0) (\theta - \theta_0) + \text{h.o.t},
\label{eq:taylor_err}
\end{align}
where $\mathbf{J}_\theta(\theta_0)$ and $\mathbf{H}_\theta(\theta_0)$ are the Jacobian and Hessian of $E$ evaluated at $\theta_0$, respectively.
Assume that the CNN is fine-tuned on \textit{old} data $\mathcal{T}_1$, which implies that $\theta_0$ would be at a local minimum (denoted by $\theta_{\mathcal{T}_1}^*$) and that $E^\prime(\theta_{\mathcal{T}_1}^*)=0$, we can rewrite Equation~\ref{eq:taylor_err} as
%
%
\begin{equation}
E_{\mathcal{T}_1}(\theta) \approx E(\theta_{\mathcal{T}_1}^*) + \frac{1}{2!} (\theta - \theta_{\mathcal{T}_1}^*)^T \mathbf{H}_\theta(\theta_{\mathcal{T}_1}^*) (\theta - \theta_{\mathcal{T}_1}^*) \\
\label{eq:approx_err}
\end{equation}
where $E_{\mathcal{T}_1}$ is the approximated error for \textit{old} data $\mathcal{T}_1$.
Next, to fine-tune the CNN on \textit{recent} data $\mathcal{T}_2$, we can use the following loss function:
\begin{equation}
E(\theta) = E_{\mathcal{T}_2}(\theta) + \beta E_{\mathcal{T}_1}(\theta),\label{eq:loss_t2}
\end{equation}
where $\beta$ controls the weighting between the two loss terms.
Substituting Equation~\ref{eq:approx_err} into Equation~\ref{eq:loss_t2}, we get
\begin{align}
E(\theta) &= E_{\mathcal{T}_2}(\theta) + \beta \Big( E(\theta_{\mathcal{T}_1}^*) + \frac{1}{2} (\theta - \theta_{\mathcal{T}_1}^*)^T \mathbf{H}_\theta(\theta_{\mathcal{T}_1}^*) (\theta - \theta_{\mathcal{T}_1}^*) \Big) \nonumber \\
&= E_{\mathcal{T}_2}(\theta) + \frac{\beta}{2} \Big( (\theta - \theta_{\mathcal{T}_1}^*)^T \mathbf{H}_\theta(\theta_{\mathcal{T}_1}^*) (\theta - \theta_{\mathcal{T}_1}^*) \Big),
\end{align}
in which the first term of Equation~\ref{eq:approx_err} is being treated as a constant and can be eliminated in the optimization.
To measure the importance of the parameters contained in the Hessian matrix, Kirkpatrick et al. approximate the Gaussian distribution of the posterior with Laplace approximation, whereby diagonal of the Fisher information matrix replaces the Hessian matrix for the approximation of the posterior~\cite{overcomeforgetting}:
\begin{equation}
E(\theta) = E_{\mathcal{T}_2}(\theta) + \frac{\beta}{2} \Big( \sum_i \mathbf{F}_i (\theta_i - \theta_{\mathcal{T}_1, i}^*)^2 \Big).
\end{equation}

\subsection{SI and MAS regularizations} \label{subsec:si_mas_reg}

Similar to EWC regularization, SI and MAS can be expressed using the following loss function:
\begin{equation}
\mathcal{L} = \mathcal{L}_{\textup{train}}  + \lambda \sum_i \boldsymbol{\Omega}_i (\theta_i - \theta_{\mathcal{T}_1, i}^*)^2, \label{eq:si_mas_reg}
\end{equation}
with the main difference being the estimation of the weight importance matrix, $ \boldsymbol{\Omega}_i $.
For MAS, the weight importance matrix is given by the gradient of the squared $L_2$-norm of the CNN output function, $ G $~\cite{mas}:
\begin{equation}
\boldsymbol{\Omega}_i = \frac{1}{N} \sum_{j=1}^{N} \frac{\partial \left\| G(x_j; \theta) \right\|^2_2}{\partial \theta_i},
\end{equation}
whereas for SI, the weight importance matrix is given by~\cite{sicontinuallearn}:
\begin{equation}
\boldsymbol{\Omega}_i = \sum_j \frac{\omega_{ij}}{(\Delta \theta_{ij})^2 + \xi},
\end{equation}
where $\omega_{ij}$ is the estimated per-parameter contribution to the total loss, and $\Delta \theta_{ij}$ the total \textit{trajectory} of the parameter.

\vspace{11pt}

\begin{IEEEbiography}[{\includegraphics[width=1in,height=1.25in,clip,keepaspectratio]{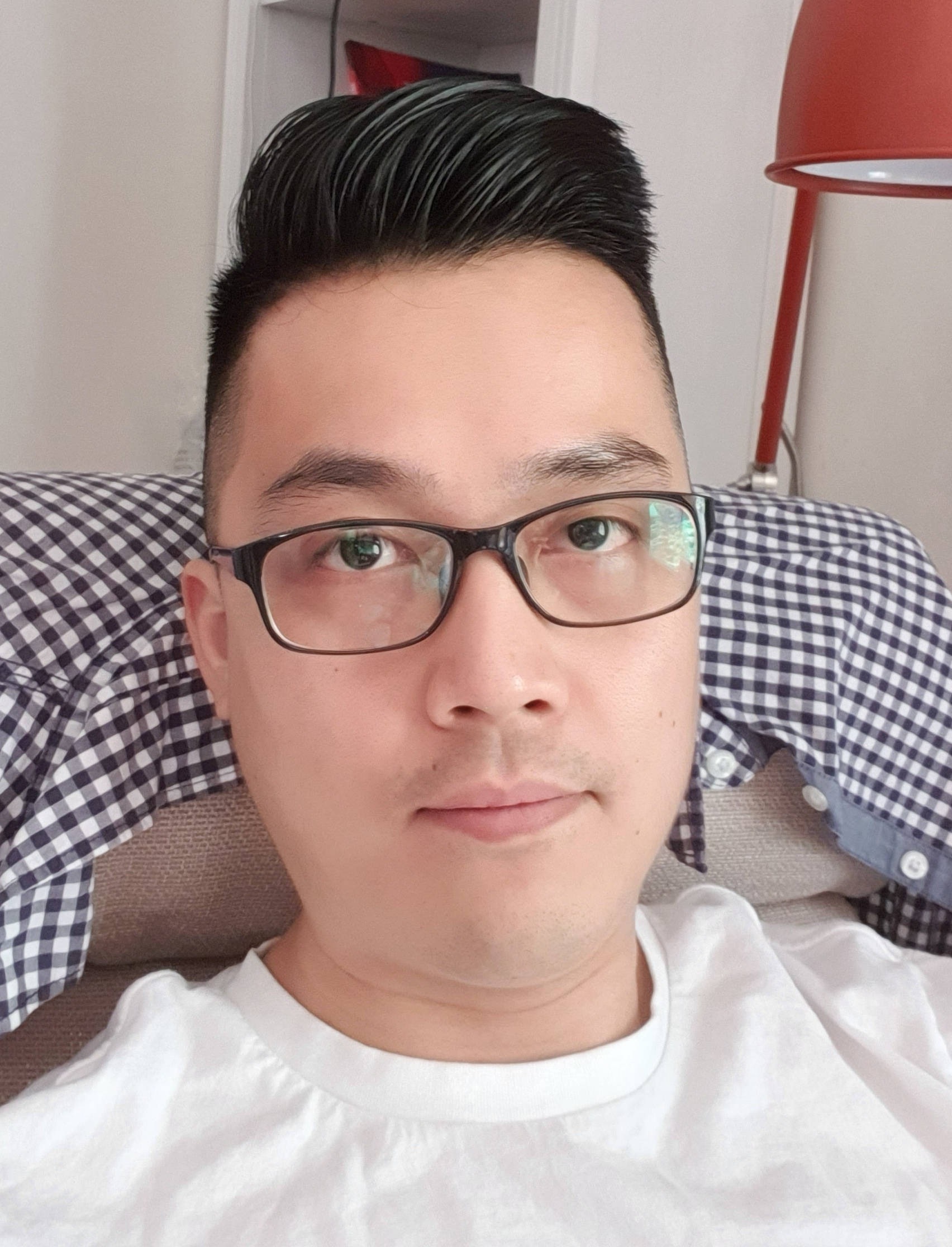}}]{Shing Yan Loo}
received his MSc degree from the Universiti Putra Malaysia (UPM). He is currently enrolled on a Dual PhD program offered by UPM and University of Alberta. His research interest lies at the intersection of visual SLAM and deep learning.
\end{IEEEbiography}

\vspace{11pt}

\begin{IEEEbiography}[{\includegraphics[width=1in,height=1.25in,clip,keepaspectratio]{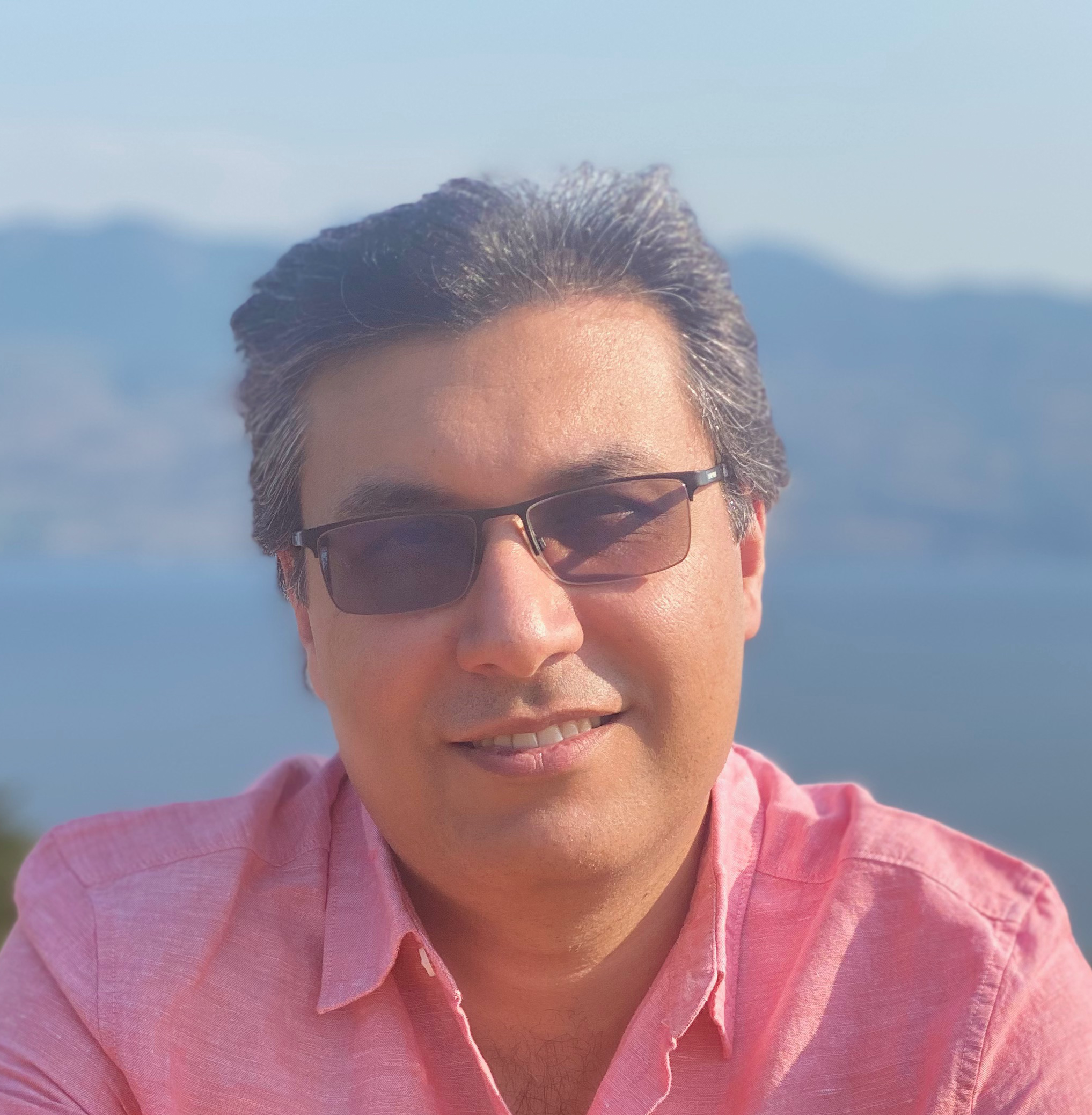}}]{Moein Shakeri}
	received the M.S. degree in computer engineering from Ferdowsi University, Iran in 2009, and Ph.D. degree in computing science from University of Alberta, Canada, 2019. During his Ph.D. he has been awarded several grants, and Ph.D outstanding thesis award. He is currently working as post-doctoral research fellow in the department of computing science at the University of Alberta. His research interests include dense map reconstruction, loop closure detection, object detection and segmentation, and polarimetric 3D reconstruction.
\end{IEEEbiography}

\vspace{11pt}

\begin{IEEEbiography}[{\includegraphics[width=1in,height=1.25in,clip,keepaspectratio]{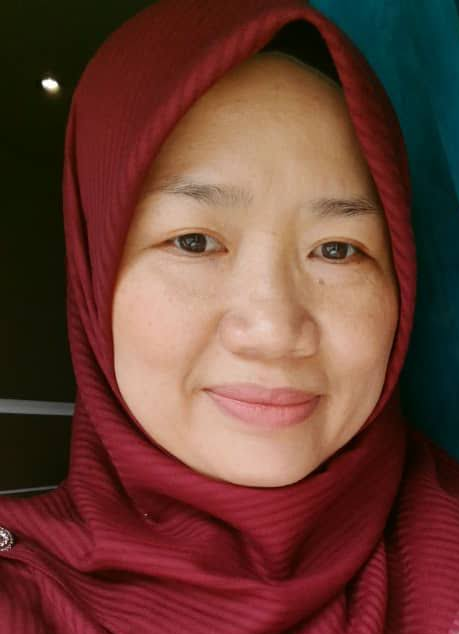}}]{S. Mashohor}
	received her B.Eng degree from the Department of Computer and Communication Systems Engineering, Universiti Putra Malaysia (UPM) in 2002. Then, she continued her study for PhD on “Genetic Algorithm based Printed Circuit Board optical inspection” at the Department of Electronics and Electrical Engineering, University of Edinburgh, UK and graduated in 2006. She has been with the Department of Computer and Communication Systems Engineering, UPM as a tutor in 2002, lecturer in 2006, senior lecturer in 2009 until 2019 and currently she is an associate professor. Her field of research is on Artificial Intelligence (AI) and Image Processing. Most of her publications are published in various AI related journals and she has been awarded with several local research grants in Malaysia. She is actively involved with multi discipline research studying AI applications in medical imaging, language and agriculture. 
\end{IEEEbiography}

\begin{IEEEbiography}[{\includegraphics[width=1in,height=1.25in,clip,keepaspectratio]{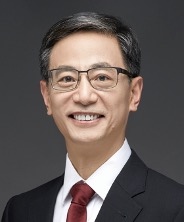}}]{Professor Hong Zhang}
	 received his B.S. from Northeastern University (Boston) in 1982, and Ph.D. from Purdue University in 1986, both in Electrical Engineering. Subsequently he conducted post-doctoral research at the University of Pennsylvania before he joined the Department of Computing Science at the University of Alberta, Canada, where he worked for over 30 years.  Since October 2020, he has been a Chair Professor in the Department of Electronic and Electrical Engineering at the Southern University of Science and Technology in Shenzhen, China.

\end{IEEEbiography}

\vspace{11pt}

\vfill

\end{document}